\documentclass[a4paper,twoside]{article}
\def\BLIND{0}
\usepackage{balance}
\usepackage{epsfig}
\usepackage{subcaption}
\usepackage{calc}
\usepackage{multirow}
\usepackage{graphicx}
\usepackage{textcomp}
\usepackage{xcolor}
\usepackage[hidelinks]{hyperref}
\usepackage[export]{adjustbox}
\usepackage{bm}
\usepackage{amssymb}
\usepackage{amstext}
\usepackage{amsmath}
\usepackage{amsthm}
\usepackage{multicol}
\usepackage{pifont}
\usepackage{pslatex}
\usepackage{caption}
\usepackage{apalike}
\usepackage{algorithm2e}
\usepackage[bottom]{footmisc}
\usepackage{SCITEPRESS}     

\begin{document}

\title{DGDNN: Decoupled Graph Diffusion Neural Network for Stock Movement Prediction}

\ifnum\BLIND=0
\author{\authorname{Zinuo You\sup{1}, Zijian Shi\sup{1}, Hongbo Bo\sup{2}, John Cartlidge\sup{1}, Li Zhang\sup{3}, Yan Ge\sup{1}}
\affiliation{\sup{1} University of Bristol}
\affiliation{\sup{2} Newcastle University}
\affiliation{\sup{3} Huawei Technologies Co., Ltd}
\email{\{zinuo.you, zijian.shi, john.cartlidge, yan.ge\}@bristol.ac.uk, hongbo.bo@newcastle.ac.uk, zhangli391@huawei.com}
}
\else
    \author{Anonymous Authors \\ 
    Anonymous Institutions \\ 
    Anonymous Contacts }
\fi

\abstract{Forecasting future stock trends remains challenging for academia and industry due to stochastic inter-stock dynamics and hierarchical intra-stock dynamics influencing stock prices. In recent years, graph neural networks have achieved remarkable performance in this problem by formulating multiple stocks as graph-structured data. However, most of these approaches rely on artificially defined factors to construct static stock graphs, which fail to capture the intrinsic interdependencies between stocks that rapidly evolve. In addition, these methods often ignore the hierarchical features of the stocks and lose distinctive information within. In this work, we propose a novel graph learning approach implemented without expert knowledge to address these issues. First, our approach automatically constructs dynamic stock graphs by entropy-driven edge generation from a signal processing perspective. Then, we further learn task-optimal dependencies between stocks via a generalized graph diffusion process on constructed stock graphs. Last, a decoupled representation learning scheme is adopted to capture distinctive hierarchical intra-stock features. Experimental results demonstrate substantial improvements over state-of-the-art baselines on real-world datasets. Moreover, the ablation study and sensitivity study further illustrate the effectiveness of the proposed method in modeling the time-evolving inter-stock and intra-stock dynamics.}


\onecolumn \maketitle \normalsize \setcounter{footnote}{0} \vfill

\section{\uppercase{Introduction}}
\label{sec:introduction}

The stock market has long been an intensively discussed research topic by investors pursuing profitable trading opportunities and policymakers attempting to gain market insights. Recent research advancements have primarily concentrated on exploring the potential of deep learning models, driven by their ability to model complex non-linear relationships~\cite{bo2023will} and automatically extract high-level features from raw data~\cite{akita2016deep,shi2022state}. These abilities further enable the capture of intricate patterns in stock market data that traditional statistical methods might omit. However, the efficient market theory~\cite{malkiel2003efficient} and the random walk nature of stock prices make it challenging to predict exact future prices with high accuracy~\cite{adam2016stock}. As a result, research efforts have shifted towards the more robust task of anticipating stock movements~\cite{jiang2021applications}.  

Early works~\cite{roondiwala2017predicting,bao2017deep} commonly adopt deep learning techniques to extract temporal features from historical stock data and predict stock movements accordingly. However, these methods assume independence between stocks, neglecting their rich connections. In reality, stocks are often interrelated from which valuable information can be derived. These complicated relations between stocks are crucial for understanding the stock markets~\cite{deng2019knowledge,feng2019temporal,feng2022relation}. 

To bridge this gap, some deep learning models attempt to model the interconnections between stocks by integrating textual data~\cite{sawhney2020deep}, such as tweets~\cite{xu2018stock} and news~\cite{li2020multimodal}. Nevertheless, these models heavily rely on the quality of embedded extra information, resulting in highly volatile performance. Meanwhile, the transformer-based methods introduce different attention mechanisms to capture inter-stock relations based on multiple time series (i.e., time series of stock indicators, such as open price, close price, highest price, lowest price, and trading volume)~\cite{yoo2021accurate,ding2021hierarchical}. Despite this advancement, these methods often lack explicit modeling of temporal information of these time series, such as temporal order and inter-series information~\cite{zhu2021survey,wen2022transformers}.

Recently, Graph Neural Networks (GNNs) have shown promising performance in analyzing various real-world networks or systems by formulating them as graph-structured data, such as transaction networks~\cite{pareja2020evolvegcn}, traffic networks~\cite{wang2020traffic}, and communication networks~\cite{li2020graph}. Typically, these networks possess multiple entities interacting over time, and time series data can characterize each entity. Analyzing stock markets as complex networks is a natural choice, as previous works indicate~\cite{liu2019risk,shahzad2018global}. Moreover, various interactive mechanisms (e.g., transmitters and receivers~\cite{shahzad2018global}) that exist between stocks can be easily represented by edges~\cite{cont2000herd}. Therefore, these properties make GNNs powerful candidates for explicitly grasping inter-stock relations and capturing intra-stock patterns with stock graphs~\cite{sawhney2021stock,xiang2022temporal}.

However, existing GNN-based models face two fundamental challenges for stock movement prediction: representing complicated time-evolving inter-stock dependencies and capturing hierarchical features of stocks. First, specific groups of related stocks are affected by various factors, which change stochastically over time~\cite{huynh2023efficient}. Most graph-based models~\cite{kim2019hats,ye2021multi,sawhney2021exploring} construct time-invariant stock graphs, which are contrary to the stochastic and time-evolving nature of the stock market~\cite{adam2016stock}. For instance, inter-stock relations are commonly pre-determined by sector or firm-specific relationships (e.g., belonging to the same industry \cite{sawhney2021exploring} or sharing the same CEO \cite{kim2019hats}). Besides, artificially defined graphs for specific tasks may not be versatile or applicable to other tasks. Sticking to rigid graphs risks introducing noise and task-irrelevant patterns to models~\cite{chen2020measuring}. Therefore, generating appropriate stock graphs and learning task-relevant topology remains a preliminary yet critical part of GNN-based methods in predicting stock movements. Second, stocks possess distinctive hierarchical features~\cite{mantegna1999hierarchical,sawhney2021exploring} that remain under-exploited (e.g., overall market trends, group-specific dynamics, and individual trading patterns~\cite{huynh2023efficient}). Previous works indicate that these hierarchical intra-stock features could distinguish highly related stocks from different levels and be utilized for learning better and more robust representations~\cite{huynh2023efficient,mantegna1999hierarchical}. However, in the conventional GNN-based methods, representation learning is combined with the message-passing process between immediate neighbors in the Euclidean space. As a result, node representations become overly similar as the message passes, severely distorting the distinctive individual node information~\cite{huang2020tackling,rusch2023survey,liu2020towards}. Hence, preserving these hierarchical intra-stock features is necessary for GNN-based methods in predicting stock movements.

In this paper, we propose the Decoupled Graph Diffusion Neural Network (DGDNN) to address the abovementioned challenges. Overall, we treat stock movement prediction as a temporal node classification task and optimize the model toward identifying movements (classes) of stocks (nodes) on the next trading day. The main contributions of this paper are summarised as follows:

\begin{itemize}
    \item We exploit the information entropy of nodes as their pair-wise connectivities with ratios of node energy as weights, enabling the modeling of intrinsic time-varying relations between stocks from the view of information propagation.
    \item  We extend the layer-wise update rule of conventional GNNs to a decoupled graph diffusion process. This allows for learning the task-optimal graph topology and capturing the hierarchical features of multiple stocks.
    \item We conduct extensive experiments on real-world stock datasets with 2,893 stocks from three markets (NASDAQ, NYSE, and SSE). The experimental results demonstrate that DGDNN significantly outperforms state-of-the-art baselines in predicting the next trading day movement, with improvements of 9.06\% in classification accuracy, 0.09 in Matthew correlation coefficient, and 0.06 in F1-Score.
\end{itemize}

\section{RELATED WORK}\label{2}
\noindent
This section provides a brief overview of relevant studies. 

\subsection{GNN-based Methods for Modeling Multiple Stocks}
\noindent
The major advantage of applying GNNs lies in their graphical structure, which allows for explicitly modeling the relations between entities. For instance, STHAN-SR~\cite{sawhney2021stock}, which is similar to the Graph Attention Neural Networks (GATs)~\cite{velivckovic2018graph}, adopts a spatial-temporal attention mechanism on a hypergraph with industry and corporate edges to capture inter-stock relations on the temporal domain and spatial domain. HATS~\cite{kim2019hats} predicts the stock movement by a GAT-based method that the immediate neighbor nodes are selectively aggregated with learned weights on manually crafted multi-relational stock graphs. Moreover, HyperStockGAT~\cite{sawhney2021exploring} leverages graph learning in hyperbolic space to capture the heterogeneity of node degree and hierarchical nature of stocks on an industry-related stock graph. This method illustrates that the node degree of stock graphs is not evenly distributed. Nonetheless, these methods directly correlate the stocks by empirical assumptions or expert knowledge to construct static stock graphs, contradicting the time-varying nature of the stock market.

\subsection{Graph Topology Learning} 
\noindent
To address the constraint of GNNs relying on the quality of raw graphs, researchers have proposed graph structure learning to optimize raw graphs for improved performance in downstream tasks. These methods can be broadly categorized into direct parameterizing approaches and neural network approaches. In the former category, methods treat the adjacency matrix of the target graph as free parameters to learn. Pro-GNN, for instance, demonstrates that refined graphs can gain robustness by learning perturbed raw graphs guided by critical properties of raw graphs~\cite{jin2020graph}. GLNN~\cite{gao2020exploring} integrates sparsity, feature smoothness, and initial connectivity into an objective function to obtain target graphs. In contrast, neural network approaches employ more complex neural networks to model edge weights based on node features and representations. For example, SLCNN utilizes two types of convolutional neural networks to learn the graph structure at both the global and local levels~\cite{zhang2020spatio}. GLCN integrates graph learning and convolutional neural networks to discover the optimal graph structure that best serves downstream tasks~\cite{jiang2019semi}. Despite these advancements, direct parameterizing approaches often necessitate complex and time-consuming alternating optimizations or bi-level optimizations, and neural network approaches may overlook the unique characteristics of graph data or lose the positional information of nodes.  

\begin{figure*}[tbhp]
\centering
  \includegraphics[width=\linewidth]{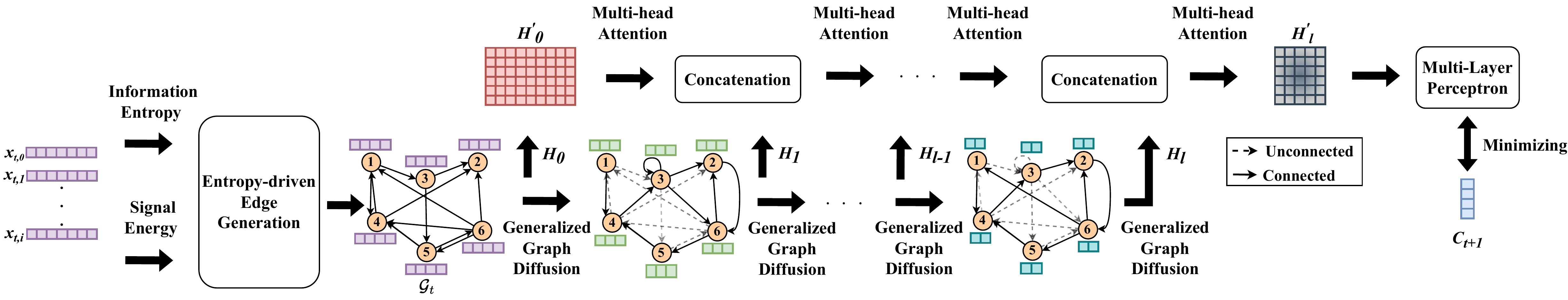}
\caption{The DGDNN framework consists of three steps: (1) constructing the raw stock graph $\mathcal{G}_t$ (see Section~\ref{C}); (2) learning the task-optimal graph topology by generalized graph diffusion (see Section~\ref{D}); (3) applying a hierarchical decoupled representation learning scheme (see Section~\ref{E}).}
  \label{fig3}
\end{figure*}  

\subsection{Decoupled Representation Learning}
\noindent
Various networks or systems exhibit unique characteristics that are challenging to capture within the constraints of Euclidean space, particularly when relying on manually assumed prior knowledge~\cite{huynh2023efficient,sawhney2021exploring}. In addressing this challenge, DAGNN~\cite{liu2020towards} offers theoretical insights, emphasizing that the entanglement between representation transformation and message propagation can hinder the performance of message-passing GNNs. SHADOW-GNN~\cite{zeng2021decoupling}, on the other hand, concentrates on decoupling the representation learning process both in depth and scope. By learning on multiple subgraphs with arbitrary depth, SHADOW-GNN preserves the distinctive information of localized subgraphs instead of globally smoothing them into white noise. Another approach, MMP~\cite{chen2022memory}, transforms updated node messages into self-embedded representations. It then selectively aggregates these representations to form the final graph representation, deviating from the direct use of representations from the message-passing process. 

\section{\uppercase{Preliminary}}\label{Preliminary}
\noindent
In this section, we present the fundamental notations used throughout this paper and details of the problem setting. Nodes represent stocks, node features represent their historical stock indicators, and edges represent interconnections between stocks.

\subsection{Notation}
\noindent
Let $\mathcal{G}_t(\mathcal{V},\mathcal{E}_{t})$ represents a weighted and directed graph on trading day $t$, where $\mathcal{V}$ is the set of nodes (stocks) $\{ v_1,...,v_N\}$ with the number of nodes as $|\mathcal{V}|=N$, and $\mathcal{E}_{t}$ is the set of edges (inter-stock relations). Let $\mathbf{A}_t \in \mathbb{R}^{N\times N}$ represents the adjacency matrix and its entry $(\mathbf{A}_t)_{i,j}$ represents an edge from $v_i$ to $v_j$. The node feature matrix is denoted as $\mathbf{X}_t\in \mathbb{R}^{N\times (\tau M)}$, where $M$ represents the number of stock indicators (i.e., open price, close price, highest price, lowest price, trading volume, etc), and $\tau$ represents the length of the historical lookback window. The feature vector of $v_i$ on trading day $t$ is denoted as $\mathbf{x}_{t,i}$. Let $c_{t,i}$ represent the label of $v_i$ on trading day $t$, where $C_t \in \mathbb{R}^{N\times 1}$ is the label matrix on trading day $t$.

\subsection{Problem Setting}
\noindent
Since we are predicting future trends of multiple stocks by utilizing their corresponding stock indicators, we transform the regression task of predicting exact stock prices into a temporal node classification task. Similar to previous works on stock movement prediction~\cite{kim2019hats,xiang2022temporal,sawhney2021stock,xu2018stock,li2021modeling}, we refer to this common and important task as next trading day stock trend classification. Given a set of stocks on the trading day $t$, the model learns from a historical lookback window of length $\tau$ (i.e., $[t-\tau+1,t]$) and predicts their labels in the next timestamp (i.e., trading day $t+1$). The mapping relationship of this work is expressed as follows,
\begin{equation}
    f(\mathcal{G}_t(\mathcal{V},\mathcal{E}_{t})) \xrightarrow{}C_{t+1}.
    \label{eqmp}
\end{equation}
Here, $f(\cdot)$ represents the proposed method DGDNN.

\section{\uppercase{Methodology}}\label{3}
\noindent
In this section, we detail the framework of the proposed DGDNN in depth, as depicted in Fig~\ref{fig3}.

\subsection{Entropy-Driven Edge Generation}\label{C}
\noindent
Defining the graph structure is crucial for achieving reasonable performance for GNN-based approaches. In terms of stock graphs, traditional methods often establish static relations between stocks through human labeling or natural language processing techniques. However, recent practices have proven that generating dynamic relations based on historical stock indicators is more effective~\cite{li2021modeling,xiang2022temporal}. These indicators, as suggested by previous financial studies~\cite{dessaint2019noisy,cont2000herd,liu2019risk}, can be treated as noisy temporal signals. Simultaneously, stocks can be viewed as transmitters or receivers of information signals, influencing other stocks~\cite{shahzad2018global,ferrer2018time}. Additionally, stock markets exhibit significant node-degree heterogeneity, with highly influential stocks having relatively large node degrees~\cite{sawhney2021exploring,arora2006financial}. 

Consequently, we propose to model interdependencies between stocks by treating the stock market as a communication network. Prior research~\cite{yue2020information} generates the asymmetric inter-stock relations based on transfer entropy. Nonetheless, the complex estimation process of transfer entropy and the limited consideration of edge weights hamper the approximation of the intrinsic inter-stock connections. 

To this end, we quantify the links between nodes by utilizing the information entropy as the directional connectivity and signal energy as its intensity. On the one hand, if the information can propagate between entities within real-world systems, the uncertainty or randomness is reduced, resulting in a decrease in entropy and an increase in predictability at the receiving entities~\cite{jaynes1957information,csiszar2004information}. On the other hand, the energy of the signals reflects their intensity during propagation, which can influence the received information at the receiver. The entry $(\mathbf{A}_t)_{i,j}$ is defined by,
\begin{equation}
    (\mathbf{A}_t)_{i,j}= \frac{E(\mathbf{x}_{t,i})}{E(\mathbf{x}_{t,j})}(e^{S(\mathbf{x}_{t,i})+S(\mathbf{x}_{t,j})-S(\mathbf{x}_{t,i}, \mathbf{x}_{t,j})}-1).
\end{equation}
\noindent
Here, $E(\cdot)$ denotes the signal energy, and $S(\cdot)$ denotes the information entropy. The signal energy of $v_i$ is obtained by,

\begin{equation}
    E(\mathbf{x}_{t,i}) = \sum_{{n=0}}^{\tau M-1} { |\mathbf{x}_{t,i}[n]|}^2.
\end{equation}
\noindent
The information entropy of $v_i$ is obtained by,

\begin{equation}
    S(\mathbf{x}_{t,i})=-\sum_{j=0}p(s_j)\ln{p(s_j)},
\end{equation}
\noindent
where $\{s_0,...,s_j\}$ denotes the non-repeating sequence of $\mathbf{x}_{t,i}$ and $p(s_j)$ represents the probability of value $s_j$. By definition, we can obtain $p(s_j)$ by,

\begin{equation}
    p(s_j)=\frac{\sum_{n=0}^{\tau M -1}\delta(s_j - \mathbf{x}_{t,i}[n])}{\tau M}.
\end{equation}
\noindent
Here $\delta(\cdot)$ denotes the Dirac delta function. 

\subsection{Generalized Graph Diffusion}\label{D}
\noindent
However, simply assuming constructed graphs are perfect for performing specific tasks can lead to discordance between given graphs and task objectives, resulting in sub-optimal model performance~\cite{chen2020measuring}. Several methods have been proposed to mitigate this issue, including AdaEdge~\cite{chen2020measuring} and DropEdge~\cite{rong2019dropedge}. These methods demonstrate notable improvements in node classification tasks by adding or removing edges to perturb graph topologies, enabling models to capture and leverage critical topological information. 

With this in mind, we propose to utilize a generalized diffusion process on the constructed stock graph to learn the task-optimal topology. It enables more effective capture of long-range dependencies and global information on the graph by diffusing information across larger neighborhoods~\cite{klicpera2019diffusion}. 

The following equation defines the generalized graph diffusion at layer $l$,

\begin{equation}
\mathbf{Q}_l=\sum_{k=0}^{K-1}\theta_{l,k}\mathbf{T}_{l,k},\;\sum_{k=0}^{K-1}\theta_{l,k}=1.
\label{eq3}
\end{equation}
\noindent
Here $\mathbf{Q}_l$ denotes the diffusion matrix, $K$ denotes the maximum diffusion step, $\theta_{l,k}$ denotes the weight coefficients, and $\mathbf{T}_{l,k}$ denotes the column-stochastic transition matrix. Specifically, generalized graph diffusion transforms the given graph structure into a new one while keeping node signals neither amplified nor reduced. Consequently, the generalized graph diffusion turns the information exchange solely between adjacent connected nodes into broader unconnected areas of the graph.

Notably, $\theta_{l,k}$ and $\mathbf{T}_{l,k}$ can be determined in advance~\cite{klicpera2019diffusion}. For instance, we can use the heat kernel or the personalized PageRank to define $\theta_{l,k}$, and the random walk transition matrix or symmetric transition matrix to define $\mathbf{T}_{l,k}$. Although these pre-defined mappings perform well in some datasets (e.g., CORA, CiteSeer, and PubMed) with time-invariant relations~\cite{zhao2021adaptive}, they are not feasible for tasks that require considering changing relationships.

Therefore, we make $\theta_{l,k}$ as trainable parameters, $\mathbf{T}_{l,k}$ as trainable matrices, and $K$ as a hyperparameter. Furthermore, we introduce a neighborhood radius~\cite{zhao2021adaptive} to control the effectiveness of the generalized graph diffusion. The neighborhood radius at layer $l$ is expressed as,

\begin{equation}
r_l=\frac{\sum_{k=0}^{K-1} \theta_{l,k}k }{\sum_{k=0}^{K-1} \theta_{l,k}}, \; r_l > 0
\label{eqr}
\end{equation}

\noindent
Here, large $r_l$ indicates the model explores more on distant nodes and vice versa.

\subsection{Hierarchical Decoupled Representation Learning}\label{E}
\noindent
Theoretically, GNNs update nodes by continuously aggregating direct one-hop neighbors, producing the final representation. However, this can lead to a high distortion of the learned representation. It is probably because the message-passing and representation transformation do not essentially share a fixed neighborhood in the Euclidean space~\cite{liu2020towards,xu2018representation,chen2020measuring}. To address this issue, decoupled GNNs have been proposed~\cite{liu2020towards,xu2018representation}, aiming to decouple these two processes and prevent the loss of distinctive local information in learned representation. Similarly, methods such as HyperStockGAT~\cite{sawhney2021exploring} have explored learning graph representations in hyperbolic spaces with attention mechanisms to capture temporal features of stocks at different levels. 
\begin{figure}[tb]
    \centering
    \includegraphics[width=0.8\linewidth]{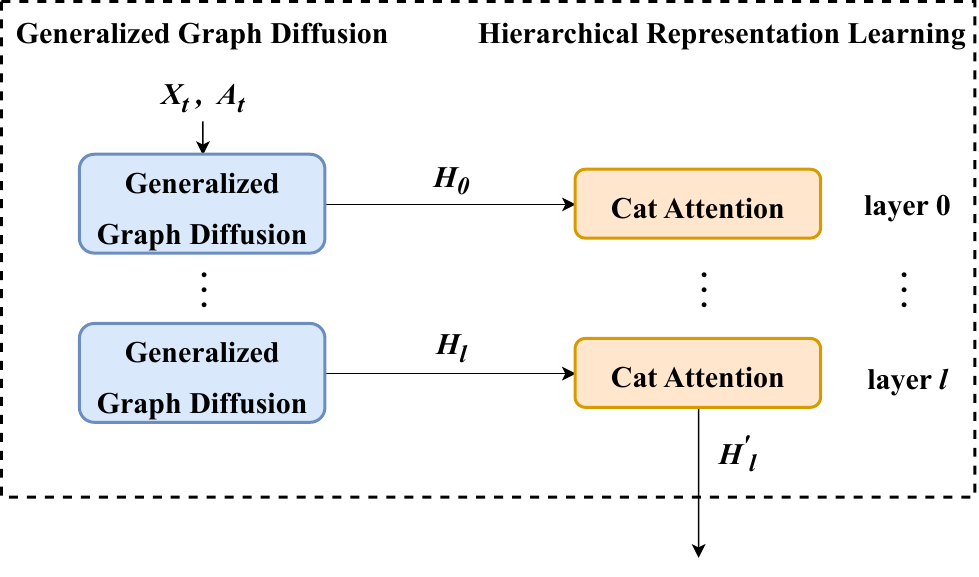}
    \caption{The component-wise layout of hierarchical decoupled representation learning with input $\mathbf{X}_t$, $\mathbf{A}_t$.}
    \label{fighfb}
\end{figure}

Inspired by these methods, we adopt a hierarchical decoupled representation learning strategy to capture hierarchical intra-stock features. Each layer in DGDNN comprises a Generalized Graph Diffusion layer and a Cat Attention layer in parallel, as depicted in Fig.~\ref{fighfb}. The layer-wise update rule is defined by,
\begin{align}
&\mathbf{H}_{l}=\sigma\left((\mathbf{Q}_l \odot \mathbf{A}_t)\mathbf{H}_{l-1}\mathbf{W}^0_l\right),\notag\\
&\mathbf{H}'_{l}=\sigma\left(\zeta(\mathbf{H}_l||\mathbf{H}'_{l-1})\mathbf{W}^1_{l}+b^1_{l}\right).
\label{eq8}
\end{align}

\noindent
Here, $\mathbf{H}'_{l}$ denote the node representation of $l-{th}$ layer, $\sigma(\cdot)$ is the activation function, $\zeta(\cdot)$ denotes the multi-head attention, $||$ denotes the concatenation, and $\mathbf{W}_{l}$ denotes the layer-wise trainable weight matrix.

\subsection{Objective Function}
\noindent
According to Eq.~\ref{eqmp}, Eq.~\ref{eq3}, and Eq.~\ref{eqr}, we formulated the objective function of DGDNN as follows,

\begin{align}
    \mathcal{J}&=\frac{1}{B} \sum_{t=0}^{B-1}\mathcal{L}_{CE}(C_{t+1}, f(\mathbf{X}_t,\mathbf{A}_t))-\alpha\sum_{l=0}^{L-1}r_l\notag\\&~~~~+\sum_{l=0}^{L-1}(\sum_{k=0}^{K-1}\theta_{l,k}-1).
\end{align}
\noindent
Here, $\mathcal{L}_{CE}(\cdot)$ denotes the cross-entropy loss, $B$ denotes the batch size, $L$ denotes the number of information propagation layers, and $\alpha$ denotes the weight coefficient controlling the neighborhood radius.

\section{EXPERIMENT}\label{4}
\noindent
The experiments are conducted on 3x Nvidia Tesla T4, CUDA version 11.2. Datasets and source code are available\footnote{\url{https://github.com/pixelhero98/DGDNN}}.

\subsection{Dataset}
\noindent 
Following previous works~\cite{kim2019hats,xiang2022temporal,sawhney2021stock,li2021modeling}, we evaluate DGDNN on three real-world datasets from two US stock markets (NASDAQ and NYSE) and one China stock market (SSE). We collect historical stock indicators from Yahoo Finance and Google Finance for all the selected stocks. We choose the stocks that span the S\&P 500 and NASDAQ composite indices for the NASDAQ dataset. We select the stocks that span the Dow Jones Industrial Average, S\&P 500, and NYSE composite indices for the NYSE dataset. We choose the stocks that compose the SSE 180 for the SSE dataset. The details of the three datasets are presented in Table~\ref{tabdata}.

\subsection{Model Setting}\label{1123}
\noindent
 Based on grid search, hyperparameters are selected using sensitivity analysis over the validation period (see Section~\ref{G}). For NASDAQ, we set $\alpha=2.9\times10^{-3}$, $\tau=19$, $K=9$, and $L=8$. For NYSE, we set $\alpha=2.7\times10^{-3}$, $\tau=22$, $K=10$, and $L=9$. For SSE, we set $\alpha=8.6\times10^{-3}$, $\tau=14$, $K=3$, and $L=5$. The training epoch is set to 1200. Adam is the optimizer with a learning rate of $2\times10^{-4}$ and a weight decay of $1.5\times10^{-5}$. The number of layers of Muti-Layer Perceptron is set to 3, the number of heads of Cat Attention layers is set to 3, the embedding dimension is set to 128,  and full batch training is selected. 
\subsection{Baseline}
\noindent
To evaluate the performance of the proposed model, we compared DGDNN with the following baseline approaches:
\begin{table}[h]
\renewcommand\arraystretch{1.5}
    \centering
    \caption{Statistics of NASDAQ, NYSE, and SSE.}
    \resizebox{\linewidth}{!}{%
    \begin{tabular}{c| c c c}
    \hline
        & NASDAQ & NYSE & SSE\\
        \hline
        Train Period & 05/2016-06/2017 & 05/2016-06/2017 & 05/2016-06/2017\\
        Validation Period & 07/2017-12/2017 & 07/2017-12/2017 & 07/2017-12/2017\\
        Test Period & 01/2018-12/2019 & 01/2018-12/2019 & 01/2018-12/2019\\
        \# Days Tr:Val:Test & 252:64:489 & 252:64:489 & 299:128:503\\
        \# Stocks & 1026 & 1737 & 130\\
        \# Stock Indicators & 5 & 5 & 4\\
        \# Label per trading day & 2 & 2 & 2\\
        \hline       
    \end{tabular}%
    }
    \label{tabdata}
\end{table}
\subsubsection{RNN-based Baseline}
\begin{itemize}
\item DA-RNN~\cite{qin2017dual}: A dual-stage attention-based RNN model with an encoder-decoder structure. The encoder utilizes an attention mechanism to extract the input time-series feature, and the decoder utilizes a temporal attention mechanism to capture the long-range temporal relationships among the encoded series.

\item Adv-ALSTM~\cite{feng2019enhancing}: An LSTM-based model that leverages adversarial training to improve the generalization ability of the stochasticity of price data and a temporal attention mechanism to capture the long-term dependencies in the price data.
\end{itemize}
\begin{table*}[t]
\renewcommand\arraystretch{1.5}
\caption{ACC, MCC, and F1-Score of proposed DGDNN and other baselines on next trading day stock trend classification over the test period. Bold numbers denote the best results.}
\centering
\resizebox{\linewidth}{!}{%
\begin{tabular}{c| c c c| c c c| c c c}
  \hline
   \multirow{2}{*}{Method} & \multicolumn{3}{c|}{NASDAQ} & \multicolumn{3}{c|}{NYSE} & \multicolumn{3}{c}{SSE}\\

   & ACC(\%) & MCC & F1-Score  & ACC(\%)  & MCC & F1-Score & ACC(\%)  & MCC & F1-Score\\ 
\hline
  DA-RNN~\cite{qin2017dual} & 57.59$\pm$0.36 & 0.05$\pm$1.47$\times10^{-3}$ & 0.56$\pm$0.01 & 56.97$\pm$0.13 & 0.06$\pm$1.12$\times10^{-3}$ & 0.57$\pm$0.02 & 56.19$\pm$0.23 & 0.04$\pm$1.24$\times10^{-3}$ & 0.52$\pm$0.02 \\
  Adv-ALSTM~\cite{feng2019enhancing} & 51.16$\pm$0.42 & 0.04$\pm$3.88$\times10^{-3}$ & 0.53$\pm$0.02 & 53.42$\pm$0.30  & 0.05$\pm$2.30$\times10^{-3}$ & 0.53$\pm$0.02 & 52.41$\pm$0.56 & 0.03$\pm$6.01$\times10^{-3}$ & 0.51$\pm$0.01 \\
  HMG-TF~\cite{ding2021hierarchical} & 57.18$\pm$0.17 & 0.11$\pm$1.64$\times10^{-3}$ & 0.59$\pm$0.01 & 58.49$\pm$0.12  & 0.09$\pm$2.03$\times10^{-3}$ & 0.59$\pm$0.02 & 58.88$\pm$0.20 & 0.12$\pm$1.71$\times10^{-3}$ & 0.59$\pm$0.01 \\
  DTML~\cite{yoo2021accurate} & 58.27$\pm$0.79 & 0.07$\pm$2.75$\times10^{-3}$ & 0.58$\pm$0.01 & 59.17$\pm$0.25 & 0.07$\pm$3.07$\times10^{-3}$ & 0.60$\pm$0.01 & 59.25$\pm$0.38 & 0.11$\pm$4.79$\times10^{-3}$ & 0.59$\pm$0.02\\
  HATS~\cite{kim2019hats} & 51.43$\pm$0.49 & 0.01$\pm$5.66$\times10^{-3}$ & 0.48$\pm$0.01 & 52.05$\pm$0.82 & 0.02$\pm$7.42$\times10^{-3}$ & 0.50$\pm$0.03 & 53.72$\pm$0.59 & 0.02$\pm$3.80$\times10^{-3}$ & 0.49$\pm$0.01\\
  STHAN-SR~\cite{sawhney2021stock} & 55.18$\pm$0.34  & 0.03$\pm$4.11$\times10^{-3}$ & 0.56$\pm$0.01 & 54.24$\pm$0.50 & 0.01$\pm$5.73$\times10^{-3}$ & 0.58$\pm$0.02 & 55.01$\pm$0.11 & 0.03$\pm$3.09$\times10^{-3}$ & 0.57$\pm$0.01\\
  GraphWaveNet~\cite{wu2019graph} & 59.57$\pm$0.27  & 0.07$\pm$2.12$\times10^{-3}$ & 0.60$\pm$0.02 & 58.11$\pm$0.66 & 0.05$\pm$2.21$\times10^{-3}$ & 0.59$\pm$0.02 & 60.78$\pm$0.23 & 0.06$\pm$1.93$\times10^{-3}$ & 0.57$\pm$0.01 \\
  HyperStockGAT~\cite{sawhney2021exploring} & 58.23$\pm$0.68  & 0.06$\pm$1.23$\times10^{-3}$ & 0.59$\pm$0.02  & 59.34$\pm$0.19  & 0.04$\pm$5.73$\times10^{-3}$ & 0.61$\pm$0.02 & 57.36$\pm$0.10  & 0.09$\pm$1.21$\times10^{-3}$ & 0.58$\pm$0.02 \\
  \hline
  DGDNN & \textbf{65.07$\pm$0.25}  & \textbf{0.20$\pm$2.33$\times10^{-3}$} & \textbf{0.63$\pm$0.01}  & \textbf{66.16$\pm$0.14}  & \textbf{0.14$\pm$1.67$\times10^{-3}$} & \textbf{0.65$\pm$0.01} & \textbf{64.30$\pm$0.32}  & \textbf{0.19$\pm$4.33$\times10^{-3}$} & \textbf{0.64$\pm$0.02}  \\
  \hline

  \end{tabular}%
  }
\label{tab1}
\end{table*}
\subsubsection{Transformer-based Baseline}
\begin{itemize}
  \item HMG-TF~\cite{ding2021hierarchical}: A transformer method for modeling long-term dependencies of financial time series. The model proposes multi-scale Gaussian priors to enhance the locality, orthogonal regularization to avoid learning redundant heads in multi-head attention, and trading gap splitter to learn the hierarchical features of high-frequency data. 
  \item DTML~\cite{yoo2021accurate}: A multi-level context-based transformer model learns the correlations between stocks and temporal correlations in an end-to-end way. 
\end{itemize}

\subsubsection{GNN-based Baseline}
\begin{itemize}
  \item HATS~\cite{kim2019hats}: A GNN-based model with a hierarchical graph attention mechanism. It utilizes LSTM and GRU layers to extract the temporal features as the node representation, and the message-passing is achieved by selectively aggregating the representation of directly adjacent nodes according to their edge type at each level.
    \item STHAN-SR~\cite{sawhney2021stock}: A GNN-based model operated on a hypergraph with two types of hyperedges: industrial hyperedges and Wikidata corporate hyperedges. The node features are generated by temporal Hawkes attention, and weights of hyperedges are generated by hypergraph attention. The spatial hypergraph convolution achieves representation and information-spreading.
  \item GraphWaveNet~\cite{wu2019graph}: A spatial-temporal graph modeling method that captures the spatial-temporal dependencies between multiple time series by combining graph convolution with dilated casual convolution.
  \item HyperStockGAT~\cite{sawhney2021exploring}: A graph attention network utilizing the hyperbolic graph representation learning on Riemannian manifolds to predict the rankings of stocks on the next trading day based on profitability. 
\end{itemize}

\subsection{Evaluation Metric}
\noindent
Following approaches taken in previous works~\cite{kim2019hats,xiang2022temporal,deng2019knowledge,sawhney2021stock,sawhney2021exploring}, F1-Score, Matthews Correlation Coefficient (MCC), and Classification Accuracy (ACC) are utilized to evaluate the performance of the models.

\begin{figure*}[tb]
    \centering
    \begin{subfigure}[b]{0.32\linewidth}
        \includegraphics[width=\textwidth]{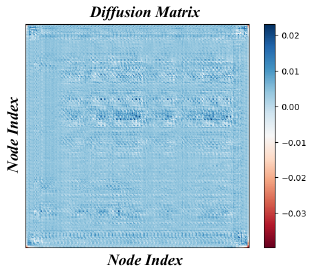}
      \caption{$\mathbf{Q}_0, t-2$}
      \label{dif1}
    \end{subfigure}
    \hfill
    \begin{subfigure}[b]{0.32\linewidth}
        \includegraphics[width=\textwidth]{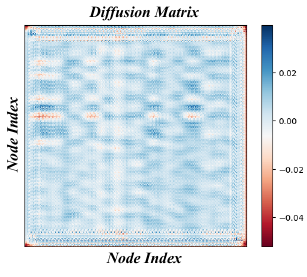}
      \caption{$\mathbf{Q}_0, t-1$}
      \label{dif2}
    \end{subfigure}
    \hfill
    \begin{subfigure}[b]{0.32\linewidth}
        \includegraphics[width=\textwidth]{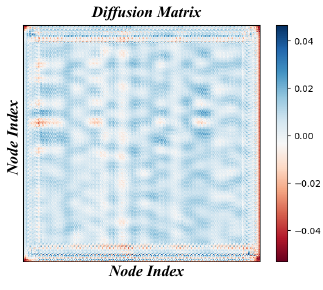}
     \caption{$\mathbf{Q}_0, t$}
     \label{dif3}
    \end{subfigure}

    \vspace{0.5cm}
    \begin{subfigure}[b]{0.32\linewidth}
        \includegraphics[width=\textwidth]{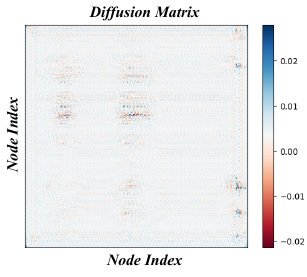}
       \caption{$\mathbf{Q}_{L-1},t-2$}
       \label{dif4}
    \end{subfigure}
    \hfill
    \begin{subfigure}[b]{0.32\linewidth}
        \includegraphics[width=\textwidth]{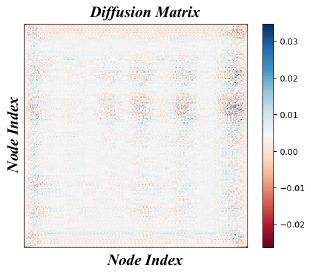}
  \caption{$\mathbf{Q}_{L-1}, t-1$}
  \label{dif5}
    \end{subfigure}
    \hfill
    \begin{subfigure}[b]{0.32\linewidth}
        \includegraphics[width=\textwidth]{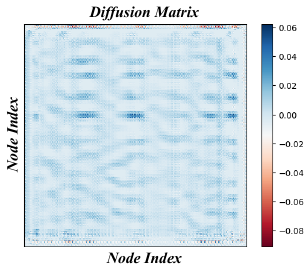}
\caption{$\mathbf{Q}_{L-1}, t$}
\label{dif6}
    \end{subfigure}
    \caption{Example normalized color maps of diffusion matrices from different layers on the NYSE dataset. $t = 03/06/2016$.}
    \label{fdif}
\end{figure*}

\subsection{Evaluation Result}
\noindent
The experimental results are presented in Table~\ref{tab1}. Our model outperforms baseline methods regarding ACC, MCC, and F1-score over three datasets. Specifically, DGDNN exhibits average improvements of 10.78\% in ACC, 0.13 in MCC, and 0.10 in F1-Score compared to RNN-based baseline methods. In comparison to Transformer-based methods, DGDNN shows average improvements of 7.78\% in ACC, 0.07 in MCC, and 0.05 in F1-Score. Furthermore, when contrasted with GNN-based models, DGDNN achieves average improvements of 7.16\% in ACC, 0.12 in MCC, and 0.07 in F1-Score.

We can make the following observations based on experimental results. First, models such as GraphWaveNet, DTML, HMG-TF, DA-RNN, and DGDNN that obtain the interdependencies between entities during the learning process perform better in most of the metrics than those methods (HATS, STHAN-SR, HyperStockGAT, and Adv-ALSTM) with pre-defined relationships (e.g., industry and corporate edges) or without considering dependencies between entities. Second, regarding the GNN-based models, HyperStockGAT and DGDNN, which learn the graph representations in different latent spaces, perform better than those (STHAN-SR and HATS) in Euclidean space. 

Fig.~\ref{fdif} presents visualizations of diffusion matrices across three consecutive trading days, with colors representing normalized weights. We make the following three observations. First, stocks from consecutive trading days do not necessarily exhibit similar patterns in terms of information diffusion. The distributions of edge weights change rapidly between Fig.~\ref{dif1} and Fig.~\ref{dif2}, and between Fig.~\ref{dif5} and Fig.~\ref{dif6}.
Second, shallow layers tend to disseminate information across a broader neighborhood. A larger number of entries in the diffusion matrices are not zero and are distributed across the matrices in Fig~\ref{dif1} to Fig.~\ref{dif3}. In contrast, deeper layers tend to focus on specific local areas. The entries with larger absolute values are more centralized in Fig.~\ref{dif4} to Fig.~\ref{dif6}). 
Third, even though the initial patterns from consecutive test trading days are similar (as shown in Fig.~\ref{dif2} and Fig.~\ref{dif3}), differences in local structures result in distinctive patterns as the layers deepen (Fig.~\ref{dif5} and Fig.~\ref{dif6}), i.e., the weights of edges can show similar distributions globally, but local areas exhibit different patterns. For instance, in Fig.~\ref{dif6}, some dark blue clusters are distinguished from light blue clusters in shape and weight, which might be crucial local graph structures.

These results suggest that the complex relationships between stocks are not static but evolve rapidly over time, and the domain knowledge does not sufficiently describe the intrinsic interdependencies between multiple entities. The manually crafted fixed stock graph assumes that the stocks of the same class are connected~\cite{livingston1977industry}, neglecting the possibility that stocks change to different classes as time changes. Besides, some stocks are more critical than others in exhibiting the hierarchical nature of intra-stock dynamics~\cite{mantegna1999hierarchical,sawhney2021exploring}, which is hard to capture in Euclidean space by directly aggregating representations as the message-passing process does.

\begin{figure*}[t]
    \centering
    \resizebox{\linewidth}{!}{
    \includegraphics[]{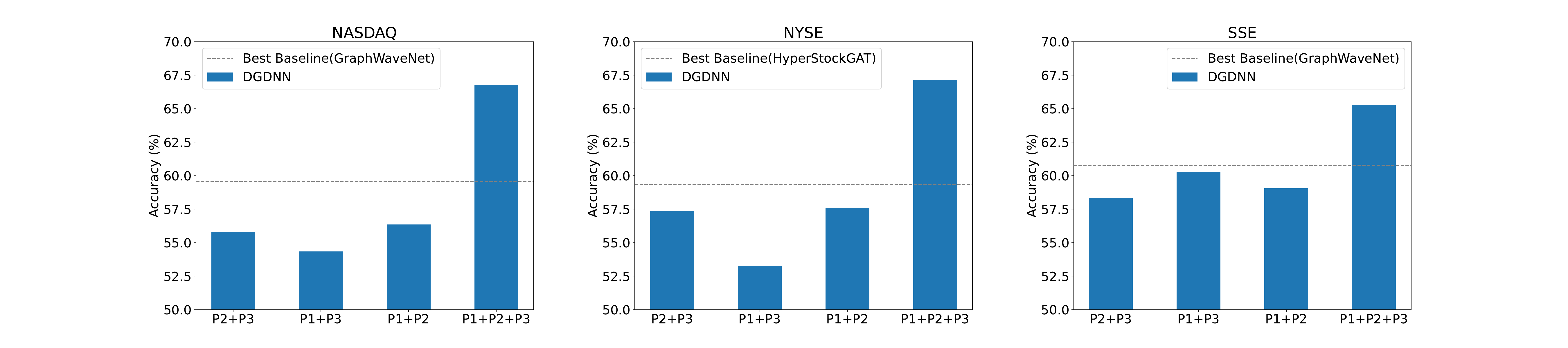}
}
\caption{Results of the ablation study. Blue: P1 denotes entropy-driven edge generation, P2 denotes generalized graph diffusion, and P3 denotes hierarchical decoupled representation learning. Gray dot line: best baseline accuracy.}
\label{fig4}
\end{figure*}

\subsection{Hyperparameter Sensitivity}\label{G}
\begin{figure}[tbhp]
   \includegraphics[width=\linewidth]{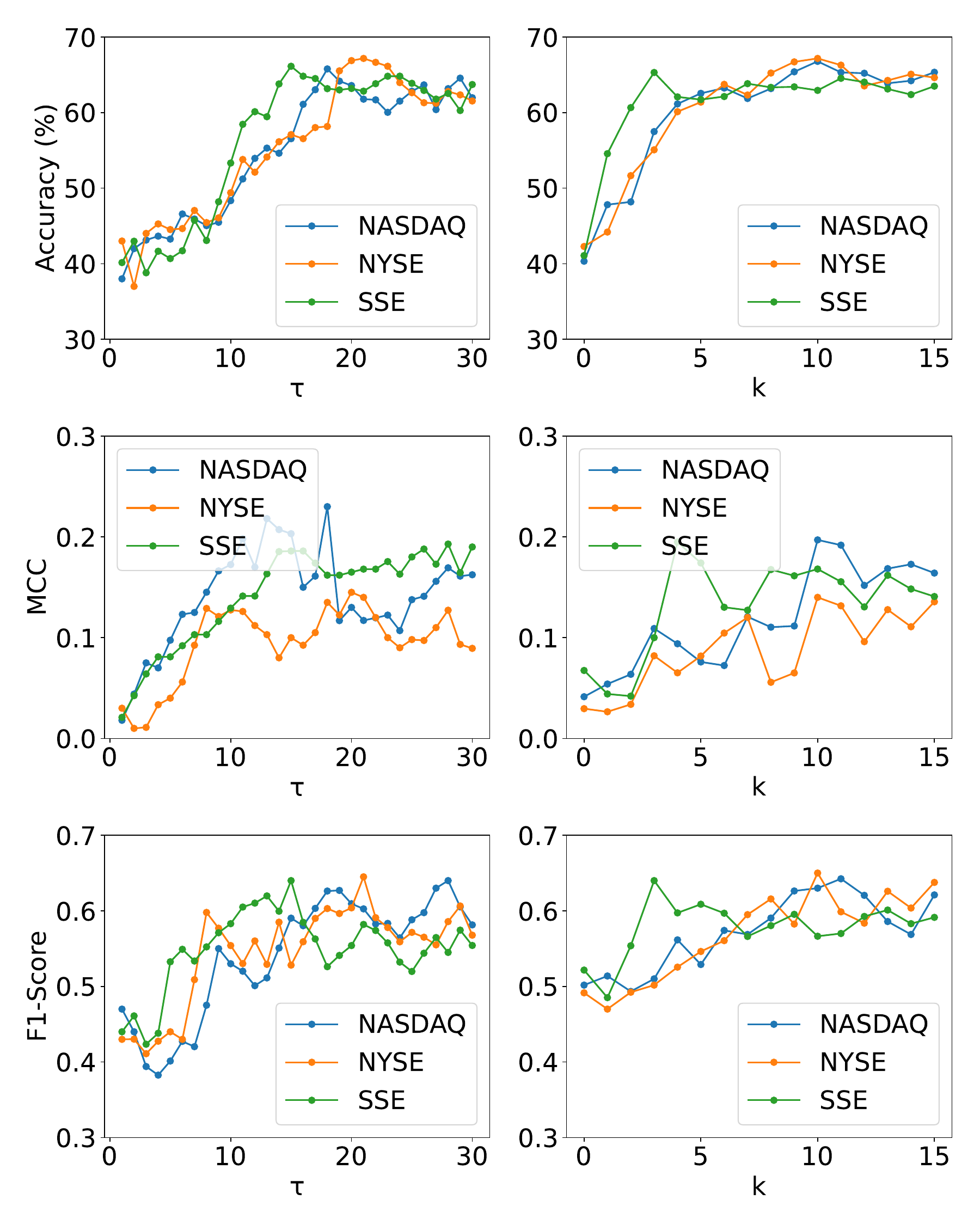}
    \caption{Sensitivity study of the historical lookback window length $\tau$ and the maximum diffusion step $K$ over validation period.}
    \label{figabcd2}
\end{figure}

\noindent
In this section, we explore the sensitivity of two important hyperparameters: the historical lookback window size $\tau$ and the maximum diffusion step $K$. These hyperparameters directly affect the model's ability to model the relations between multiple stocks. The sensitivity results of $\tau$ and $K$ are shown in Fig.~\ref{figabcd2}.
Based on the sensitivity results, DGDNN consistently performs better on the three datasets when the historical lookback window size $\tau\in [14,24]$. This coincides with the 20-day (i.e., monthly) professional financial strategies~\cite{adam2016stock}. Moreover, the optimal $K$ of DGDNN varies considerably with different datasets. On the one hand, the model's performance generally improves as $K$ grows on the NASDAQ dataset and the NYSE dataset, achieving the optimal when $K \in \{9,10\}$. On the other hand, the model's performance on the SSE dataset reaches the peak when $K=3$ and retains a slightly worse performance as $K$ grows. Intuitively, the stock graph of the SSE dataset is smaller than the NASDAQ dataset and the NYSE dataset, resulting in a smaller $K$.

\subsection{Ablation Study}
\noindent
The proposed DGDNN consists of three critical components: entropy-driven edge generation, generalized graph diffusion, and hierarchical decoupled representation learning. We further verify the effectiveness of each component by removing it from DGDNN. The ablation study results are shown in Fig.~\ref{fig4}. 

\subsubsection{Entropy-driven Edge Generation} To demonstrate the effectiveness of constructing dynamic relations from the stock signals, we replace the entropy-driven edge generation with the commonly adopted industry-corporate stock graph using Wikidata\footnote{\url{https://www.wikidata.org/wiki/Wikidata:List_of_properties}}~\cite{feng2019temporal}. We observe that applying the industry and corporate relationships leads to an average performance reduction of classification accuracy by 9.23\%, reiterating the importance of considering temporally evolving dependencies between stocks. Moreover, when testing on the NYSE dataset and the SSE dataset, the degradation of model performance is slightly smaller than on the NASDAQ dataset. According to financial studies~\cite{jiang2011comparison,schwert2002stock}, the NASDAQ market tends to be more unstable than the other two. This might indicate that the injection of expert knowledge works better in less noisy and more stable markets. 

\subsubsection{Generalized Graph Diffusion} We explore the impact of utilizing the generalized graph diffusion process. Results of the ablation study show that DGDNN performs worse without generalized graph diffusion on all datasets, with classification accuracy reduced by 10.43\% on average. This indicates that the generalized graph diffusion facilitates information exchange better than immediate neighbors with invariant structures. While the performance degradation on the SSE dataset is about 38\% of the performance degradation on the NASDAQ dataset and the NYSE dataset. Since the size of the stock graphs (130 stocks) of the SSE dataset is much smaller than the other two (1026 stocks and 1737 stocks), the graph diffusion process has limited improvements through utilizing larger neighborhoods.

\subsubsection{Hierarchical Decoupled Representation Learning} 
The ablation experiments demonstrate that the model coupling the two processes deteriorates with a reduction of classification accuracy by 9.40\% on the NASDAQ dataset, 8.55\% on the NYSE dataset, and 5.23\% on the SSE dataset. This observation empirically validates that a decoupled GNN can better capture the hierarchical characteristic of stocks. Meanwhile, this suggests that the representation transformation is not necessarily aligned with information propagation in Euclidean space. It is because different graphs exhibit various types of inter-entities patterns and intra-entities features, which do not always follow the assumption of smoothed node features~\cite{liu2020towards,xu2018representation,li2018deeper}.

\section{CONCLUSION}\label{5}
\noindent
In this paper, we propose DGDNN, a novel graph learning approach for predicting the future trends of multiple stocks based on their historical indicators. Traditionally, stock graphs are crafted based on domain knowledge (e.g., firm-specific and industrial relations) or generated by alternative information (e.g., news and reports). To make stock graphs appropriately represent complex time-variant inter-stock relations, we dynamically generate raw stock graphs from a signal processing view considering financial theories of stock markets. Then, we propose leveraging the generalized graph diffusion process to optimize the topologies of raw stock graphs. Eventually, the decoupled representation learning scheme captures and preserves the hierarchical features of stocks, which are often overlooked in prior works. The experimental results demonstrate performance improvements of the proposed DGDNN over baseline methods. The ablation study results prove the effectiveness of each module in DGDNN. Besides financial applications, the proposed method can be easily transferred to tasks that involve multiple entities exhibiting interdependent and time-evolving features. One limitation of DGDNN is that it generates an overall dynamic relationship from multiple stock indicators without sufficiently considering the interplay between them. Notwithstanding the promising results, we plan to learn multi-relational dynamic stock graphs and allow information to be further diffused across different relational stock graphs in future work.

\balance

\bibliographystyle{apalike}
{\small
\bibliography{example}}

\end{document}